%% file: main.tex
\documentclass{article}

\input{src/macro}

\title{Towards Collaborative Plan Acquisition \\ through Theory of Mind Modeling in Situated Dialogue}

\author{
    Cristian-Paul Bara\textsuperscript{\rm 1}, Ziqiao Ma\textsuperscript{\rm 1}, Yingzhuo Yu\textsuperscript{\rm 1}, Julie Shah\textsuperscript{\rm 2}, Joyce Chai\textsuperscript{\rm 1}
    \affiliations
    \textsuperscript{\rm 1}University of Michigan, \textsuperscript{\rm 2}Massachussets Institute of Technology
    \emails
    \{cpbara,marstin,yyzjason,chaijy\}@umich.edu, julie\_a\_shah@csail.mit.com
}

\begin{document}

\maketitle

\input{src/0-abstract}
\input{src/1-introduction}
\input{src/2-background}

\input{src/3-1-data}
\input{src/3-2-task}
\input{src/3-3-method}

\input{src/4-experiment}
\input{src/5-conclusion}

\clearpage

\input{src/6-acknowledge}
\bibliographystyle{named}
\bibliography{reference}

\clearpage
\appendix
\input{src/7-appendix}

\end{document}

%% file: src/macro.tex
\usepackage{ijcai23}

\usepackage{times}
\usepackage{soul}
\usepackage{url}
\usepackage[hidelinks]{hyperref}
\usepackage[utf8]{inputenc}
\usepackage[small]{caption}
\usepackage{graphicx}
\usepackage{amsmath}
\usepackage{amsthm}
\usepackage{booktabs}
\usepackage{algorithm}
\usepackage{algorithmic}
\usepackage{subcaption}
\usepackage{xcolor}
\usepackage{blindtext}

\usepackage{enumerate,enumitem}

\newtheorem{definition}{Definition}

%% file: src/0-abstract.tex
\begin{abstract}

Collaborative tasks often begin with partial task knowledge and incomplete initial plans from each partner. 
To complete these tasks, agents need to engage in situated communication with their partners and coordinate their partial plans towards a complete plan to achieve a joint task goal. 
While such collaboration seems effortless in a human-human team, it is highly challenging for human-AI collaboration. 
To address this limitation, this paper takes a step towards \textbf{collaborative plan acquisition}, where humans and agents strive to learn and communicate with each other to acquire a complete plan for joint tasks. 
Specifically, we formulate a novel problem for agents to predict the missing task knowledge for themselves and for their partners based on rich perceptual and dialogue history. 
We extend a situated dialogue benchmark for symmetric collaborative tasks in a 3D blocks world and investigate computational strategies for plan acquisition. 
Our empirical results suggest that predicting the partner's missing knowledge is a more viable approach than predicting one's own. 
We show that explicit modeling of the partner's dialogue moves and mental states produces improved and more stable results than without.
These results provide insight for future AI agents that can predict what knowledge their partner is missing and, therefore, can proactively communicate such information to help their partner acquire such missing knowledge toward a common understanding of joint tasks. 

\end{abstract}

%% file: src/1-introduction.tex
\section{Introduction}
\label{sec:introduction}

With the simultaneous progress in AI assistants and robotics, it is reasonable to anticipate the forthcoming milestone of embodied assistants. 
However, a critical question arises: how can we facilitate interactions between humans and robots to be as intuitive and seamless as possible? 
To bridge this gap, a significant challenge lies in the mismatched knowledge and skills between humans and agents, as well as their tendency to begin with incomplete and divergent partial plans.
Considering that the average user cannot be expected to possess expertise in robotics, language communications become paramount.
Consequently, it becomes imperative for humans and agents to engage in effective language communication to establish shared plans for collaborative tasks. 
While such coordination and communication occur organically between humans, it's notoriously difficult for human-AI teams, particularly when it comes to physical robots acquiring knowledge from situated interactions involving intricate language and physical activities.

To address this challenge, this paper takes a first step towards {\bf collaborative plan acquisition} (CPA), where humans and agents aim to communicate, learn, and infer a complete plan for joint tasks through situated dialogue. 
To this end, we extended \texttt{MindCraft}~\cite{bara-etal-2021-mindcraft}, a benchmark for symmetric collaborative tasks with disparate knowledge and skills in a 3D virtual world.  
Specifically, we formulate a new problem for agents to predict the absent task knowledge for themselves and for their partners based on a wealth of perceptual and dialogue history. 
We start by annotating fine-grained dialogue moves, which capture the communicative intentions between partners during collaboration. 
Our hypothesis is that understanding communicative intentions plays a crucial role in ToM modeling, which, in turn, facilitate the acquisition of collaborative plans.
We developed a sequence model that takes the interaction history as input and predicts the dialogue moves, the partner's mental states, and the complete plan. 
Our empirical results suggest that predicting the partner's missing knowledge is a more viable approach than predicting one's own. 
We show that explicit modeling of the partner's dialogue moves and mental states produces improved and more stable results than without.

The contributions of this work lie in that it bridges collaborative planning with situated dialogue to address how partners in a physical world can collaborate to arrive at a joint plan. 
In particular, it formulates a novel task on missing knowledge prediction and demonstrates that it's feasible for agents to predict their partner's missing knowledge with respect to their own partial plan. 
Our results have shown that, in human-AI collaboration, a more viable collaboration strategy is to infer and tell the partner what knowledge they might be missing and prompt the partner for their own missing knowledge. 
This strategy, if adopted by both agents, can potentially improve common ground in collaborative tasks.
Our findings will provide insight for developing embodied AI agents that can collaborate and communicate with humans in the future. 

%% file: src/2-background.tex
\section{Related Work}
\label{sec:background}

Our work bridges several research areas, particularly in the intersection of human-robot collaboration, planning, and theory of mind modeling.

\subsection{Mixed-Initiative Planning}

The essence of a collaborative task is that two participants, human or autonomous agents, pool their knowledge and skills to achieve a common goal. 
A mixed-initiative planning system involves a human planner and an automated planner with a goal to reduce the load and produce better plans~\cite{lino2005semantic}. 
We notice an implication that the agent's functional role is a supportive one. 
Work in this paradigm, called \textit{intelligent decision support}, involves agents that range from low-level processing and/or visualization~\cite{lino2005semantic} to offering higher-level suggestions~\cite{manikonda2014ai}. 
Examples on this spectrum provide agents checking constraint satisfaction, eliciting user feedback on proposed plans, and ins some cases the agents are the primary decision makers~\cite{zhang2012human,gombolay2015decision,sengupta2017radar}.
Our desire is to have human-robot collaboration starting from an equal footing and a key goal of this is to resolve disparities in starting knowledge and abilities. 
We believe this can be learned from observing human-human interaction and will lead to better quality collaboration. 
Prior work indicates that mutual understanding and plan quality can be improved between intelligent agents through interaction~\cite{kim2016improving}, though most of the results are qualitative~\cite{di2000agreement}, makes abstract implications \cite{kim2016improving}, are not tractable~\cite{grosz1996collaborative}. 
Hybrid probabilistic generative and logic-based models that overcome incomplete information and inconsistent observations have been proposed by Kim et al.~\shortcite{kim2015inferring}. 
These were successfully applied to observe natural human team planning conversations and infer the agreed-upon plan.
Following these footprints, we introduce a collaborative plan acquisition task to explicitly tackle the initial disparities in knowledge and abilities of a human planner and an automated planner.

\subsection{Goal and Plan Recognition}

Goal recognition (GR)~\cite{heinze2004modelling,moniz2013state} refers to the problem of inferring the goal of an agent based on the observed actions and/or their effects.
On top of that, plan recognition (PR)~\cite{kautz1986generalized,carberry2001techniques} further challenges AI agents to construct a complete plan by defining a structure with the set of observed and predicted actions that will lead to the goal~\cite{sukthankar2014plan,van2021activity}. 
We introduce a collaborative plan acquisition (CPA) task as a step forward along this line.
In the CPA setting, humans and agents start both with incomplete task knowledge, communicate with each other to acquire a complete plan, and actively act in a shared environment for joint tasks.
We further discuss some key benefits of our setting compared to existing work.
In terms of experiment setup, the majority of the current approaches employ plan libraries with predefined sets of possible plans~\cite{avrahami2005fast,mirsky2016sequential}, or domain theories to enable plan recognition as planning (PRP)~\cite{ramirez2009plan,sohrabi2016plan}, which suffer from scalability issue in complex domains~\cite{pereira2017landmark} for high-dimensional data~\cite{amado2018lstm}.
Motivated by existing research~\cite{rabkina2020recognizing,rabkina2021evaluation}, we adapt Minecraft as our planning domain, as it allows us to define agents with hierarchical plan structures and visual perception in a 3D block world that requires plan recognition from latent space.
In terms of task setup, the setting of the CPA task shares the merit of active goal recognition~\cite{shvo2020active}, where agents are not passive observers but are enabled to sense, reason, and act in the world. 
We further enable agents to communicate with their partners with situated dialogue, which is more realistic in real-world human-robot interaction.
Although there exists research to integrate non-verbal communication to deal with incomplete plans in sequential plan recognition~\cite{mirsky2016sequential,mirsky2018sequential} and research to integrate natural language processing through parsing~\cite{geib2007natural}, little work was done to explore language communication and dialogue processing.
The CPA task introduces a more general and symmetric setting, where agents not only query their partners for missing knowledge but also actively share knowledge that their partners may be ignorant of.

\subsection{Theory of Mind Modeling}

As introduced by Premack and Woodruff~\shortcite{premack1978chimpanzee}, one has a Theory of Mind (ToM) if they impute \textit{mental states} to themselves and others.
While interacting with others, Humans use their ToM to predict partners' future actions~\cite{dennett1988precis}, to plan to change others' beliefs and next actions~\cite{ho2022planning} and to facilitate their own decision-making~\cite{rusch2020theory}.
In recent years, the AI community has made growing efforts to model a machine ToM to strengthen agents in human-robot interaction~\cite{kramer2012human} and multiagent coordination~\cite{albrecht2018autonomous}.
We compare our work with representative work along this line in two dimensions.
In terms of the role of the agent, prior research is largely limited to passive observer roles~\cite{grant2017can,nematzadeh2018evaluating,le2019revisiting,rabinowitz2018machine} or as a speaker in a Speaker-Listener relationship~\cite{zhu2021fewshot}. 
Following a symmetric and collaborative setup~\cite{bara-etal-2021-mindcraft,sclar2022symmetric}, we study ToM modeling in agents that actively interact with the environment and engage in free-form situated communication with a human partner.
In terms of task formulation, machine ToM has been typically formulated as inferring other agents’ beliefs~\cite{grant2017can,nematzadeh2018evaluating,le2019revisiting,bara-etal-2021-mindcraft}, predicting future actions~\cite{rabinowitz2018machine}, generating pragmatic instructions~\cite{zhu2021fewshot}, and gathering information~\cite{sclar2022symmetric}.
None of these formulations were able to explicitly assess how well can AI agents use their machine ToM to complete partial plans through situation communication with their collaborators, as humans usually do in real-world interactions.
To this end, we extended the problem formulation in~\cite{bara-etal-2021-mindcraft} to a collaborative plan acquisition task, where humans and agents try to learn and communicate with each other to acquire a complete plan for joint tasks.

\begin{figure*}[htb!]
    \centering
    \includegraphics[width=0.999\textwidth]{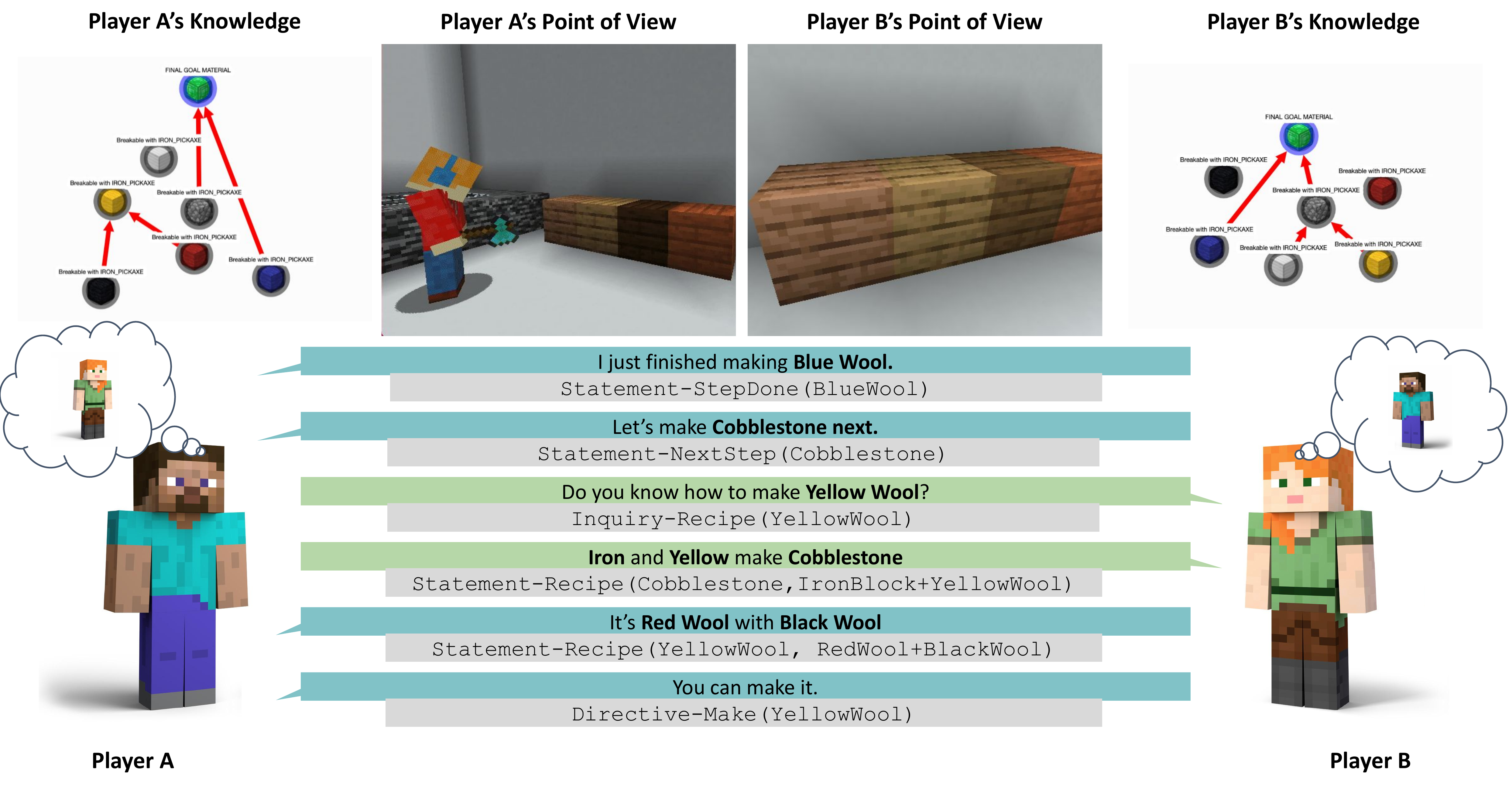}
    \vspace{-12pt}
    \caption{An example dialogue history between two partners to complete a joint goal in \texttt{MindCraft}. Each player is given a partial plan. They communicate with each other to form a complete plan for the goal. Each utterance is annotated with a dialogue move that describes the communicative intention. \vspace{-8pt}}
    \label{fig:mindcraft}
\end{figure*}

%% file: src/3-1-data.tex
\section{Background: ToM for Collaborative Tasks}

\begin{figure*}[htb!]
    \centering
    \includegraphics[width=0.9\textwidth]{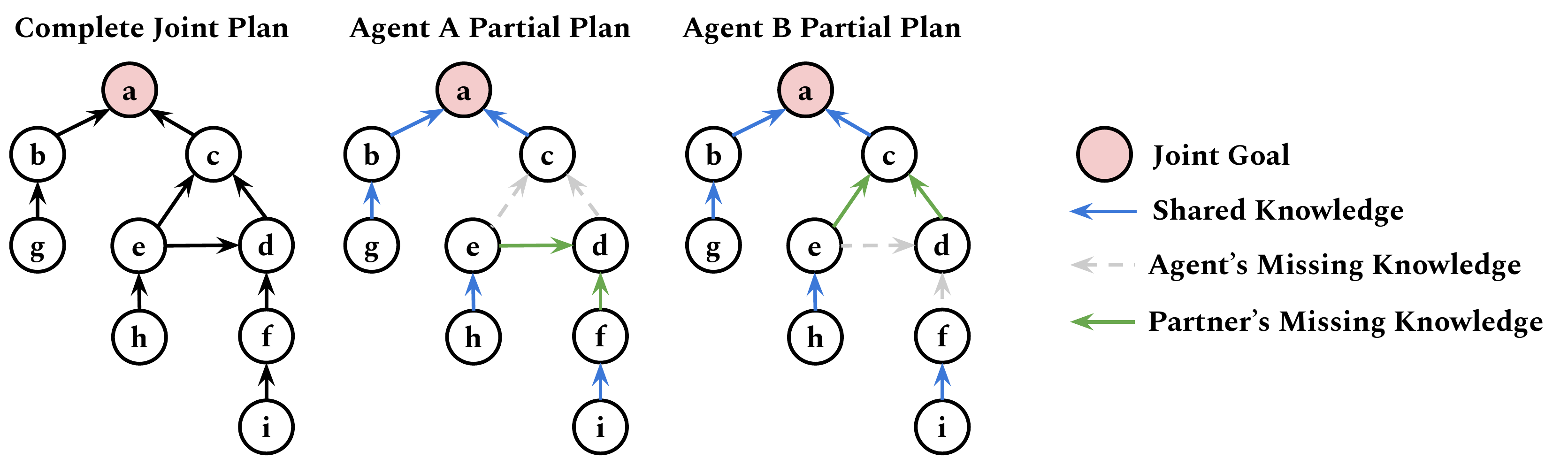}
    \vspace{-8pt}
    \caption{An example of the plan graphs. From left to right, we illustrate a complete joint plan, a partial plan for Agent A, and another partial plan for Agent B. The plan graphs all contain the same set of nodes, with a \textit{joint goal} $a$ on top and other nodes representing \textit{fact landmarks}. In a collaborative plan acquisition problem, an agent is tasked to infer its own missing knowledge and its partner's missing knowledge. \vspace{-8pt}}
    \label{fig:plan}
\end{figure*}

Our work is built upon \texttt{MindCraft}, a benchmark and platform developed by ~\cite{bara-etal-2021-mindcraft} for studying ToM modeling in collaborative tasks. 
We first give a brief introduction to this benchmark and then illustrate how our work differs from \texttt{MindCraft}.\footnote{The original dataset consists of 100 dialogue sessions. We used the platform and collected 60 additional sessions to increase the data size for our investigation under the IRB (HUM00166817) approved by the University of Michigan.}

The \texttt{MindCraft} platform supports agents to complete collaborative tasks through situated dialogue in a 3D block world, with rich mental state annotations. 
As shown in Figure~\ref{fig:mindcraft}, two agents are co-situated in a shared environment and their joint goal is to create a specific material.
There are two macro-actions: (1) creating a block and (2) combining two existing blocks to create a new block. 
These macro-actions are made up of atomic actions that agents may perform in-game, \textit{e.g.}, navigation, jumping, and moving blocks around.
During gameplay, the two agents are each given a partial plan in the form of a directed AND-graph. 
These partial plans are incomplete in the sense that the agents cannot complete the task by themselves individually by following the plan.
The two players will need to communicate with each other through the in-game chat so that they can coordinate with each other to complete the joint task.
An example dialogue session between the players to communicate the plan is shown in Figure~\ref{fig:mindcraft}.

As an initial attempt, Bara et al.~\shortcite{bara-etal-2021-mindcraft} formulated ToM modeling as three tasks that predict a partner's mental state:

\begin{itemize}[leftmargin=*]
    \setlength\itemsep{-0.25em}
    \item \textbf{Task Intention}: predict the sub-goal that the partner is currently working on;
    \item \textbf{Task Status}: predict whether the partner believes a certain sub-goal is completed and by whom;
    \item \textbf{Task Knowledge}: predict whether the partner knows how to achieve a sub-goal, \textit{i.e.}, all the incoming edges of a node;
\end{itemize}
A baseline was almost implemented that takes perceptual observation and interaction history for these prediction tasks and reported results in~\cite{bara-etal-2021-mindcraft}. 
For the remainder of this paper, we use ToM tasks to refer to these three tasks introduced in the original paper.

It's important to note that, although we use the \texttt{MindCraft} benchmark, our work here has several significant differences. 
First and foremost, while \texttt{MindCraft} studies ToM, it primarily focused on inferring other agents’ mental states, and has not touched upon collaborative plan acquisition. 
How humans and agents communicate, learn, and infer a complete plan for joint tasks through situated dialogue is the new topic we attempt to address in this paper. 
Second, \texttt{MindCraft} mostly focuses on ToM modeling and only provides ground-truth labels for the three tasks described above. 
As communicative intentions play an important role in coordinating activities between partners, we added additional annotations for dialogue moves (as shown in Figure~\ref{fig:mindcraft}).
We investigate if the incorporation of dialogue moves would benefit mental state prediction and plan acquisition.

%% file: src/3-2-task.tex
\section{Collaborative Plan Acquisition}
\label{sec:task}

In a human-AI team like that in \texttt{MindCraft}, humans and AI agents may have insufficient domain knowledge to derive a complete plan, thus suffering from an incomplete action space to execute a complete plan. 
It's therefore important for an agent to predict what knowledge is missing for themselves and for their partners, and proactively seek/share that information so the team can reach a common and complete plan for the joint goal.
We start by formalizing the plan representation, followed by a description of the collaborative plan acquisition problem.

\subsection{Task Formulation}

\begin{definition} [Joint and Partial Plan]
\label{def:coplan}
We represent a joint plan $\mathcal{P} = (V,E)$ as a directed AND-graph, where the nodes $V$ denote sub-goals and the edges $E$ denote temporal constraints between the subgoals.
As a directed AND-graph, all of the children sub-goals of an AND-node must be satisfied in order to perform the parent.
A partial plan $\tilde{\mathcal{P}} = (V, \tilde{E})$ is a subgraph of $\mathcal{P}$ with a shared set of nodes $V$ but only share a subset of edges $\tilde{E}\subseteq E$. 
\end{definition}

An example of a complete plan graph in \texttt{MindCraft} is shown in Figure~\ref{fig:plan}.
Each plan contains a joint goal, and the rest of the nodes denotes fact landmarks~\cite{hoffmann2004ordered}, \textit{i.e.}, sub-goals that must be achieved at some point along all valid execution.

We consider a pair of collaborative agents with a joint plan $\mathcal{P}$. 
To account for the limited domain knowledge, an agent $i$ has an initial partial plan $\mathcal{P}_i = (V, E_i), E_i\subseteq E$, which is a subgraph of $\mathcal{P}$.
As shown in Figure~\ref{fig:plan}, the complete plan and partial plans share the same set of nodes (\textit{i.e.}, the same $V$).
The agent only has access to its own knowledge, which might be shared (denoted as blue arrows) or not (denoted as green arrows).
Its missing knowledge (denoted as grey arrows) is a set of edges $\overline{E}_i = E\backslash E_i$.
We assume, in this work, the collaborative planning problem is solvable for the collaborative agents, \textit{i.e.}, $\bigcup E_i = E$.
In order for the agents to solve the problem, agents can communicate with their partners by sending a natural language message, which in turn helps them to acquire a complete plan graph.
We define this collaborative plan acquisition problem formally as follows.

\begin{definition} [Collaborative Plan Acquisition Problem]
\label{def:cpa}
In a collaborative plan acquisition problem with a joint plan $\mathcal{P}$, an agent $i$ and its collaborative partner $j$ start with partial plans $\mathcal{P}_i = (V, E_i)$ and $\mathcal{P}_j = (V, E_j)$. 
At each timestamp $t$, the agent $i$ has access to a sequence of up-to-date visual observations $O_i^{(t)}$ and dialogue history $D^{(t)}$.
The problem is for agent $i$'s to acquire its own missing knowledge $\overline{E}_i = E\backslash E_i$ and the partner $j$'s missing knowledge $\overline{E}_j = E\backslash E_j$.
\end{definition}

To solve a collaborative plan acquisition problem, agent $i$ needs to address two tasks.

\paragraph{Task 1: Inferring Own Missing Knowledge.}
For Task 1, a solution is a set of missing knowledge $\overline{E}_i = E\backslash E_i$. 
Agent $i$ needs to infer its own missing knowledge by identifying the missing edges in its partial plan from the complete joint plan, \textit{i.e.}, $P(e \in E \backslash E_i\ |\ O_i^{(t)}, D_i^{(t)}),\forall e\in V^2\backslash E_i$ at time $t$.
Note that $V^2$ refers to a complete graph, instead of a complete joint plan.
For example, as shown in Figure~\ref{fig:plan}, among all the missing edges in Agent A's partial plan, we hope Agent A would correctly predict that the edge $d\rightarrow c$ and the edge $e \rightarrow c$ are missing in their own plan. Recovering those edges leads to a complete joint plan.

\paragraph{Task 2: Inferring Partner's Missing Knowledge.}
For Task 2, a solution is a set of missing knowledge $\overline{E}_j = E\backslash E_j$.
Agent $i$ predicts what edges in $i$'s partial plan which might be missing in its partner $j$'s partial plan, \textit{i.e.}, $P(e \in E \backslash E_j\ |\ O_i^{(t)}, D_i^{(t)}),\forall e\in E_i$ at time $t$.
In the example in Figure~\ref{fig:plan}, Agent A should select the edges $e\rightarrow d$ and $f\rightarrow d$ from their own partial plan as being absent from their partner's plan. If the agent can correctly predict which edges are missing for their partner, the agent can proactively communicate to their partner and therefore help their partner acquire a complete task plan. If agents can predict what each other is missing and proactively share that knowledge, then both agents will be able to reach a common understanding of the complete joint plan.

We note that the \textbf{Task Knowledge} in ToM tasks is different from the \textbf{Task 2} we propose. 
In Task Knowledge, the model is probed whether \textit{one} piece of knowledge that might be unknown itself is known by the partner. 
In Task 2, the model needs to predict, for \textit{each} piece of the agent’s own knowledge, whether the partner shares it or not.

%% file: src/3-3-method.tex
\begin{figure*}[htb!]
    \centering
    \includegraphics[width=0.97\textwidth,trim={0 9cm 20cm 0},clip]{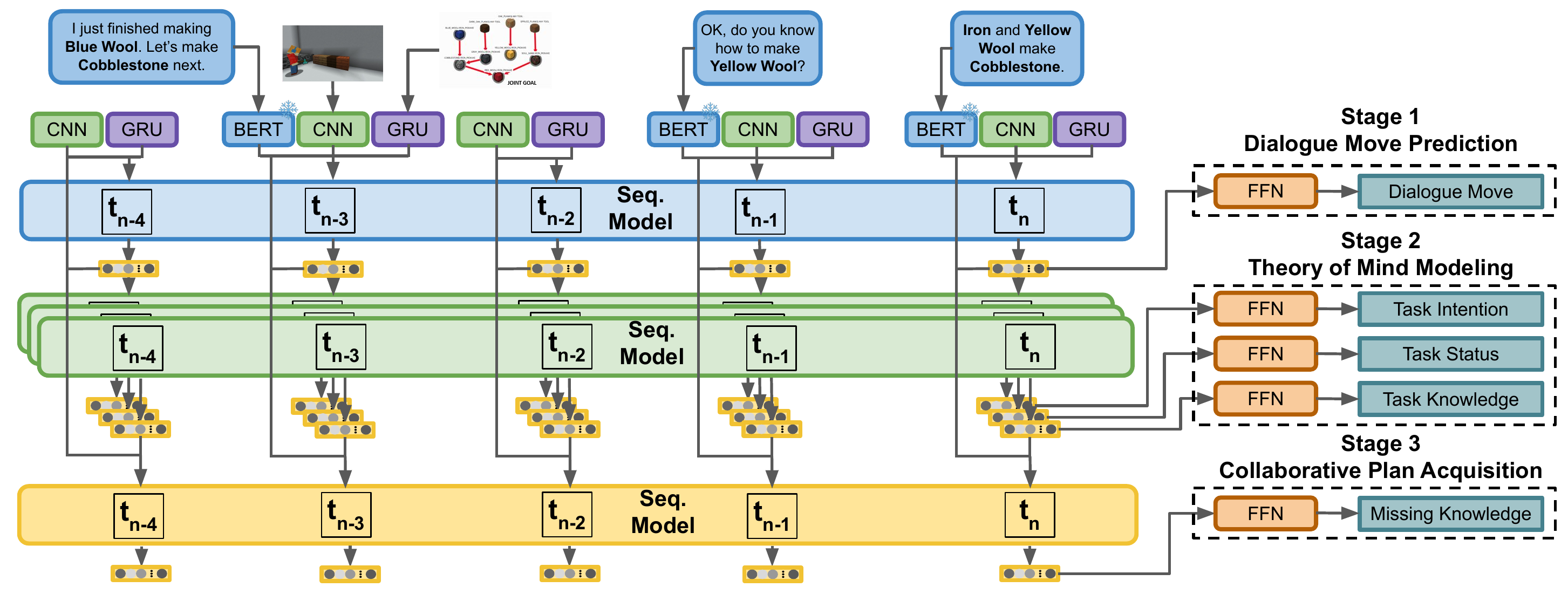}
    \vspace{-4pt}
    \caption{The theory of mind (ToM) model consists of a base sequence model taking in as input representations for dialogue ($D$) when available), visual observation of the environment ($O$), and the partial plan available to the agent. The model can be configured to take optional inputs as latent representations from the frozen mental state prediction models and the dialogue move representation for dialogue exchanges. \vspace{-8pt}}
    \label{fig:sequence_model}
\end{figure*}

\subsection{Dialogue Moves in Coordination of Plans}
\label{sec:dialogue}

While the ToM tasks have captured the partner's \textit{task intention}, another important dimension of intention, the \textit{communicative intention}, was neglected in the original \texttt{MindCraft} paper.
This can be captured by \textit{dialogue moves}, which are sub-categories of dialogue acts that guide the conversation and update the information shared between the speakers~\cite{traum2003information,marge2020let}.
To this end, we introduce a dialogue move prediction task to better understand the dialogue exchanges between the agent and its partner in a collaborative planning problem. 
We build our move schema from the set of dialogue acts described in~\cite{stolcke2000dialogue}, out of which we keep a relevant subset. 
We expand the \texttt{Directive}, \texttt{Statement}, and \texttt{Inquiry} categories for domain-specific uses in the \texttt{MindCraft} collaborative task.  
For these three categories, we introduce parameters that complete the semantics of dialogue moves.
These parameters serve the purpose of grounding the dialogue to the partial plan graph given to the player. 
We show all dialogue moves with parameters in 
Table~\ref{tab:dialogue_move_taxonomy} in Section~\ref{sec:annotation} in 
the supplemental material. 

\subsection{Computational Model}

\paragraph{End-to-End Baseline.}
We start by introducing a straightforward end-to-end approach to address both Task 1 and Task 2, similar to the model in \cite{bara-etal-2021-mindcraft} for consistency. 
More specifically, the baseline model processes the dialogue input with a frozen language model~\cite{devlin2019bert}, the video frames with a convolutional neural network, and the partial plan with a gated recurrent unit~\cite{chung2014empirical}.
The sequences of visual, dialogue, and plan representations are fed into an LSTM~\cite{hochreiter1997long}. 
The predictions of the missing knowledge of the agent and its partner are decoded with feed-forward networks.

\paragraph{Augmented Models.}
One important research question we seek to address is whether explicit treatment of the partner's mental states and dialogue moves would benefit collaborative plan acquisition.
As shown in Figure~\ref{fig:sequence_model}, we develop a multi-stage augmented model that first learns to predict the dialogue moves, then attempts to model a theory of mind, and finally learns to predict the missing knowledge of itself and of its partner respectively.
The model processes each input modality and time series with the same architectural design as the baseline.
At Stage 1, the models predict the dialogue moves of the partner.
At Stage 2, we freeze the model pre-trained in Stage 1 and concatenate their output latent representations to augment the input sequences of visual, dialogue, and plan representations. The models predict the mental states of the partner, with each LSTM sequence model dedicated to task intention, task status, and task knowledge.
At Stage 3, we freeze the models pre-trained in Stages 1 and 2 and task the model to predict the missing knowledge with similar LSTM and feed-forward networks. 
\footnote{Our code is available at \url{https://github.com/sled-group/collab-plan-acquisition}.}

%% file: src/4-experiment.tex
\section{Empirical Studies}
\label{sec:evaluation}

In this section, we first examine the role of dialogue moves in three ToM tasks and further discuss how dialogue moves and ToM modeling influence the quality of collaborative plan acquisition.

\subsection{Role of Dialogue Moves in ToM Tasks}

The performance of the mental state prediction models, as presented in \texttt{MindCraft}, shows low performance in the multimodal setting.
We begin by confirming the effectiveness of dialogue moves, by evaluating if they help to improve these ToM tasks proposed in~\cite{bara-etal-2021-mindcraft}. 
The same setting as described by the Stage 1 and 2. 
We show results in Table~\ref{tab:ToM_performance_with_dlg_moves}, which compares the performance of the baseline model with the model augmented with dialogue moves. 
We observe a significant increase in performance when using dialogue moves. 
Furthermore, for the task of predicting the partner's task knowledge, we observe that the augmented model approaches the average human performance.\footnote{The average human performance measure was provided in the original \texttt{MindCraft}~\cite{bara-etal-2021-mindcraft}.} 
The best-performing models for every ToM task are used to produce the latent representations for our subsequent tasks.

\begin{table}[htb!]
    \centering
    \resizebox{0.9\columnwidth}{!}{
        \begin{tabular}{|l|c|c|c|}
            \hline
            \multicolumn{4}{|c|}{\textbf{Task Status}} \\
            \hline
            \textbf{Modalities} & \textbf{w/o Dlg Moves} & \textbf{w/ Dlg Moves} & \textbf{Human} \\ 
            \hline
            None    &  N/A     & \textbf{56.0}$\pm0.8$    & 67.0 \\
            D       & 45.8$\pm3.0$  & 54.6$\pm1.1$  & 67.0 \\
            D+O     & 32.7$\pm1.2$  & \textbf{59.3}$\pm1.0$  & 67.0 \\
            O       & 53.7$\pm1.1$  & \textbf{59.3}$\pm1.7$  & 67.0 \\
            \hline\hline
            \multicolumn{4}{|c|}{\textbf{Task Knowledge}} \\
            \hline
            \textbf{Modalities} & \textbf{w/o Dlg Moves} & \textbf{w/ Dlg Moves} & \textbf{Human} \\ 
            \hline
            None    &    N/A    & \textbf{54.7}$\pm2.5$  & 58.0 \\
            D       & 45.3$\pm1.3$  & \textbf{56.2}$\pm1.9$  & 58.0 \\
            D+O     & 48.3$\pm1.1$  & \textbf{57.6}$\pm1.0$  & 58.0 \\
            O       & 49.4$\pm2.5$  & \textbf{56.4}$\pm2.5$  & 58.0 \\
            \hline\hline
            \multicolumn{4}{|c|}{\textbf{Task Intention}} \\
            \hline
            \textbf{Modalities} & \textbf{w/o Dlg Moves} & \textbf{w/ Dlg Moves} & \textbf{Human} \\ 
            \hline
            None    &   N/A       & \textbf{14.9}$\pm1.5$  & 46.0 \\
            D       &  3.0$\pm0.6$  & \textbf{12.1}$\pm1.0$  & 46.0 \\
            D+O     &  6.2$\pm0.6$  & \textbf{13.5}$\pm0.6$  & 46.0 \\
            O       &  6.6$\pm1.1$  & \textbf{13.8}$\pm1.7$  & 46.0 \\
            \hline
        \end{tabular}
    }
    \caption{Performance on the three ToM tasks with or without using dialogue moves. \texttt{None} means only dialogue moves are used for prediction. \texttt{D} stands for text in dialogue; \texttt{O} stands for visual observation of the activities in the environment; \texttt{D+O} both text and visual observation. Highlighted values are statistically significant with $P<0.01$ compared to the best model without dialogue moves for the given task. \vspace{-8pt}
    }
    \label{tab:ToM_performance_with_dlg_moves}
\end{table}

\subsection{Results for Collaborative Plan Acquisition}

We now present the empirical results for collaborative plan acquisition, \textit{i.e.}, inferring one's own missing knowledge and inferring the partner's missing knowledge at the end of each session. 
We use the following metrics to evaluate the performance on these tasks: 
\begin{itemize}[leftmargin=*]
    \item \textbf{Per Edge F1 Score}, computed by aggregating all edges across tasks. It is meant to evaluate the model's ability to predict whether an edge is missing in a partial plan.
    \item \textbf{Per Task F1 Score}, computed as the average of F1 scores within a dialogue session. It is meant to evaluate the model's average performance across sessions.
\end{itemize}

\paragraph{Task 1: Inferring Own Missing Knowledge.} 
The performance is shown in Table \ref{tab:complete_own_plan}. 
Overall, we found the models underperform across all configurations, meaning that inferring one's own missing knowledge turns out to be a difficult task.
We believe this is due to the sparsity of the task graph.
Since the space of possible edges to be predicted is large (as the agent needs to consider every possible link between two nodes), the link prediction becomes notoriously challenging.
Better solutions will be needed for this task in the future.

\begin{table}[!ht]
    \centering
    \resizebox{0.9\columnwidth}{!}{
        \begin{tabular}{|c|c|c|c|c|c|}
        \hline
        \textbf{Task} & \textbf{Task} & \textbf{Task}. & \textbf{Dlg}. & \textbf{Per Edge} & \textbf{Per Task} \\
        \textbf{Status} & \textbf{Know}. & \textbf{Int.} & \textbf{Move} & \textbf{F1 Score} & \textbf{F1 Score} \\
        \hline
  &       &       &       &       17.0 $\pm$   0.2  &      19.8 $\pm$   1.0        \\ 
  &  \texttt{X} &       &       &       19.0 $\pm$   2.5  &      21.1 $\pm$   1.6        \\ 
  &       &  \texttt{X} &       &       \textbf{21.0 } $\pm$   0.7  &      22.2 $\pm$   2.2        \\ 
  &  \texttt{X} &  \texttt{X} &       &       \textbf{19.6} $\pm$   1.4  &      21.4 $\pm$   1.7        \\ 
  &  \texttt{X} &       &  \texttt{X} &       \textbf{20.1} $\pm$   1.4  &      22.1 $\pm$   1.2        \\ 
  &       &  \texttt{X} &  \texttt{X} &       \textbf{19.8} $\pm$   1.7  &      21.7 $\pm$   1.8        \\ 
\texttt{X} &  \texttt{X} &  \texttt{X} &  \texttt{X} &       17.4 $\pm$   0.1  &      20.0 $\pm$   1.9        \\ 
\hline
        \end{tabular}
    }
    \vspace{-4pt}
    \caption{Performance on inferring agent's own knowledge. Highlighted values are statistically significant with $P<0.01$ compared to the base model without augmentation.\vspace{-8pt} }
    \label{tab:complete_own_plan}
\end{table}

\begin{table}[!ht]
    \centering
    \resizebox{0.9\columnwidth}{!}{
        \begin{tabular}{|c|c|c|c|c|c|}
        \hline
        \textbf{Task} & \textbf{Task} & \textbf{Task}. & \textbf{Dlg}. & \textbf{Per Edge} & \textbf{Per Task} \\
        \textbf{Status} & \textbf{Know}. & \textbf{Int.} & \textbf{Move} & \textbf{F1 Score} & \textbf{F1 Score} \\
        \hline
  &       &       &       &       71.3 $\pm$   1.1  &      68.8 $\pm$   3.1  \\ 
  &  \texttt{X} &       &       &       \textbf{74.4} $\pm$   0.3  &      73.6 $\pm$   1.4  \\ 
  &       &  \texttt{X} &       &       \textbf{74.5} $\pm$   1.4  &      73.8 $\pm$   3.0  \\ 
  &  \texttt{X} &   \texttt{X} &       &       \textbf{74.4} $\pm$   1.4  &      73.5 $\pm$   2.3  \\ 
  &  \texttt{X} &       &  \texttt{X} &       \textbf{74.3} $\pm$   0.7  &      \textbf{73.5} $\pm$   1.3  \\ 
  &       &  \texttt{X} & \texttt{X} &       \textbf{75.0} $\pm$   1.0  &      \textbf{74.7} $\pm$   2.2  \\ 
\texttt{X} &  \texttt{X} &  \texttt{X} &  \texttt{X} &       \textbf{73.5} $\pm$   0.5  &      72.1 $\pm$   1.8  \\ 
\hline
        \end{tabular}
    }
    \vspace{-4pt}
    \caption{Performance on inferring the player's partner's plan. Highlighted values are statistically significant with $P<0.01$ compared to the base model without augmentation.\vspace{-8pt}}
    \label{tab:complete_partner_plan}
\end{table}

\paragraph{Task 2: Inferring Partner's Missing Knowledge.} 
Table \ref{tab:complete_partner_plan} shows the performance which is across the board more than 70\% F1. 
This means that this task is more approachable compared to inferring one's own missing knowledge. 
Table \ref{tab:complete_partner_plan} also compares various combinations of augmented models, with different augmentations available from Stage 1 (\textit{e.g.}, latent representations of dialogue moves and other mental states). 
While all combinations lead to increased performance, we found in general that incorporating dialogue moves and task intention has the highest performance. 
This finding confirms the importance of intention prediction in plan coordination, from both the communication level and task level.

\subsection{Cross-time Analysis}

\begin{figure*}[!ht]
    \centering
    \begin{subfigure}{\columnwidth}
        \includegraphics[width=1.0\columnwidth]{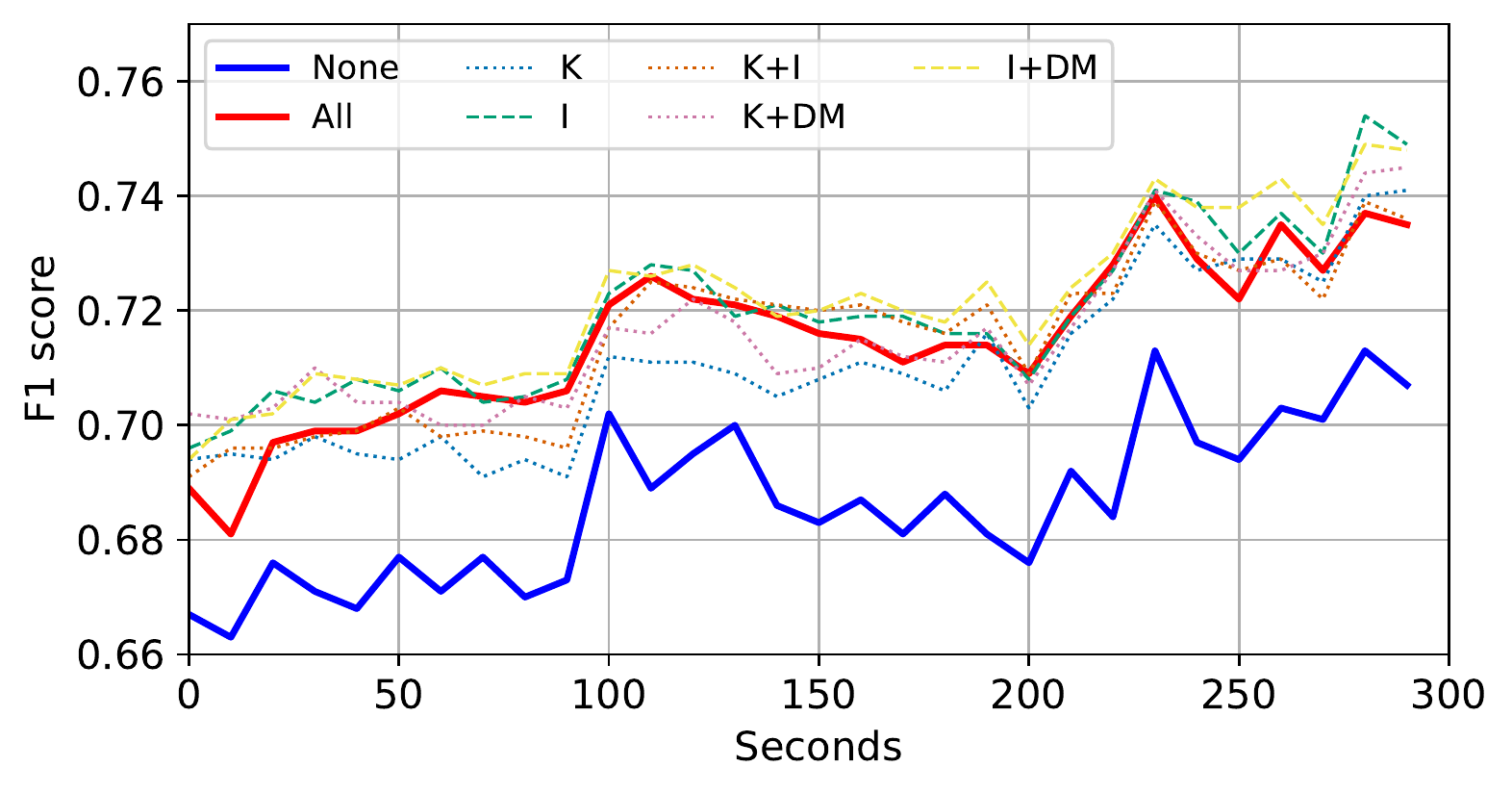}
        \vspace{-15pt}
        \caption{Per edge F1 score}
        \label{fig:cumulative_per_edge_f1_absolute_time}
    \end{subfigure}
    \begin{subfigure}{\columnwidth}
        \includegraphics[width=1.0\columnwidth]{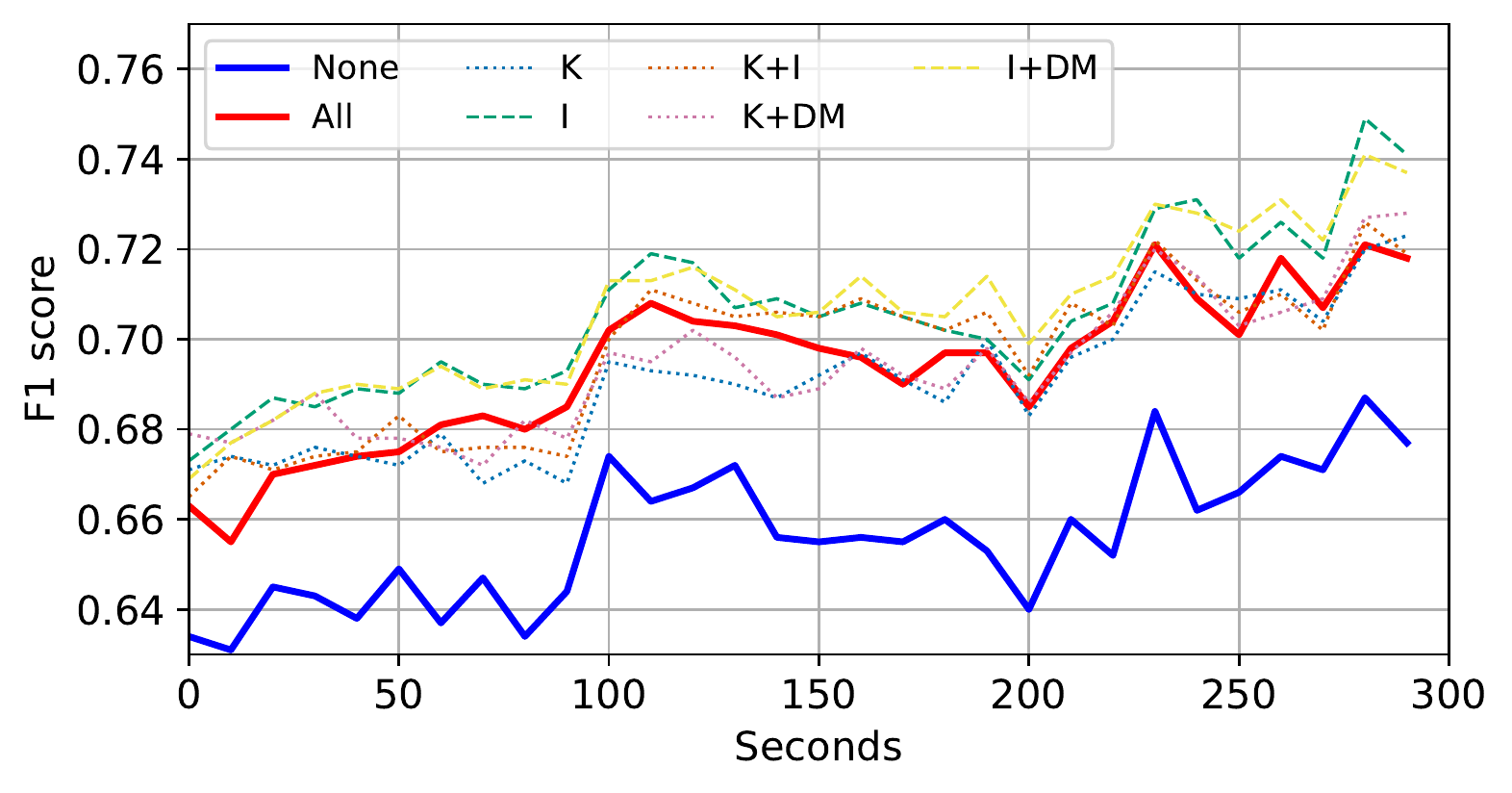}
        \vspace{-15pt}
        \caption{Per task F1 score}
        \label{fig:cumulative_per_game_f1_absolute_time}
    \end{subfigure}
    \vspace{-5pt}
    \caption{Model performance as the interaction progresses in absolute time.
    The abbreviations are S - Task Status, K - Task Knowledge, I - Task Intent, and DM - Dialogue Moves.\vspace{-3pt}}
    \label{fig:cumulative_absolute_time}
\end{figure*}

\begin{figure*}[!ht]
    \centering
    \begin{subfigure}{\columnwidth}
        \includegraphics[width=1.0\columnwidth]{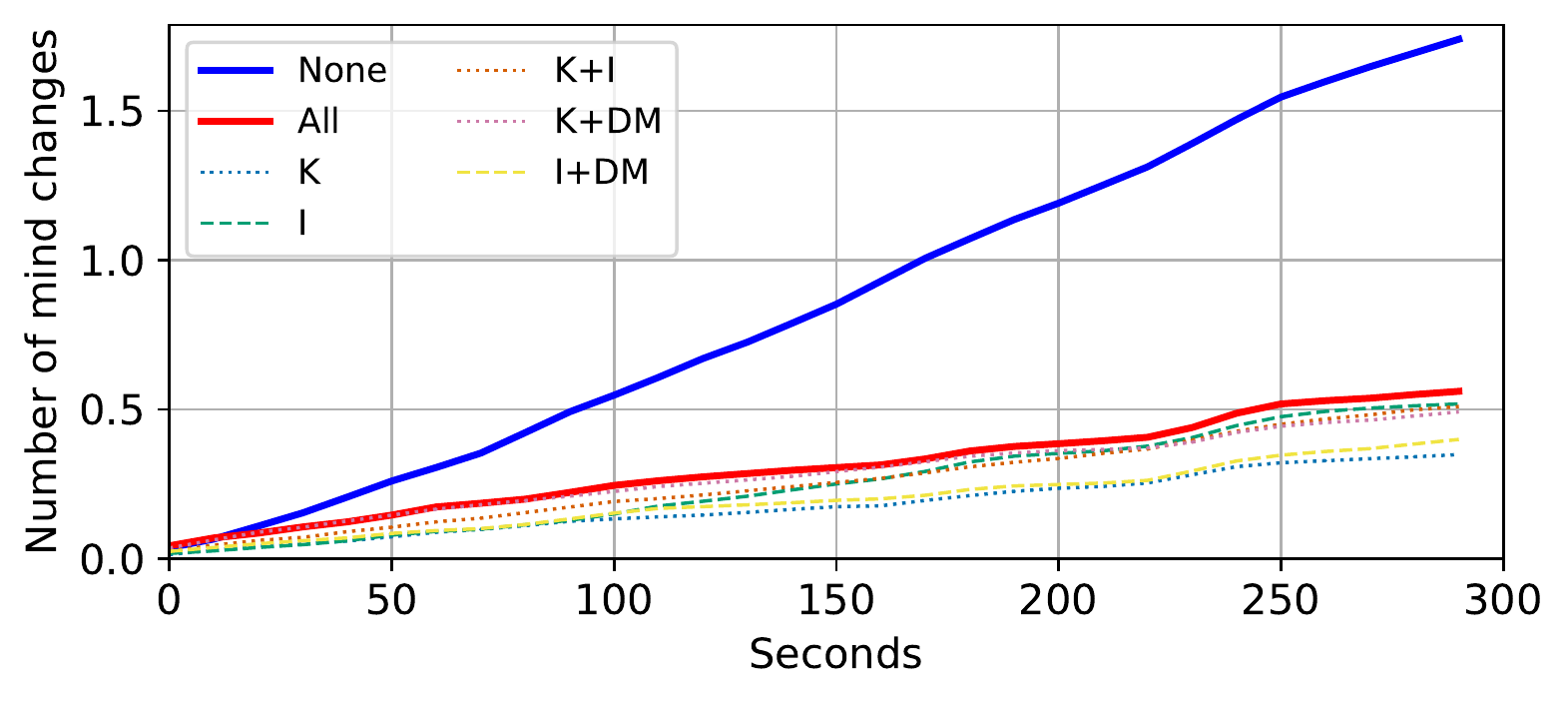}
        \vspace{-15pt}
        \caption{Cumulative number of times the model changes its mind.}
        \label{fig:cumulative_mind_changes_absolute_time}
    \end{subfigure}
    \begin{subfigure}{\columnwidth}
        \includegraphics[width=1.0\columnwidth]{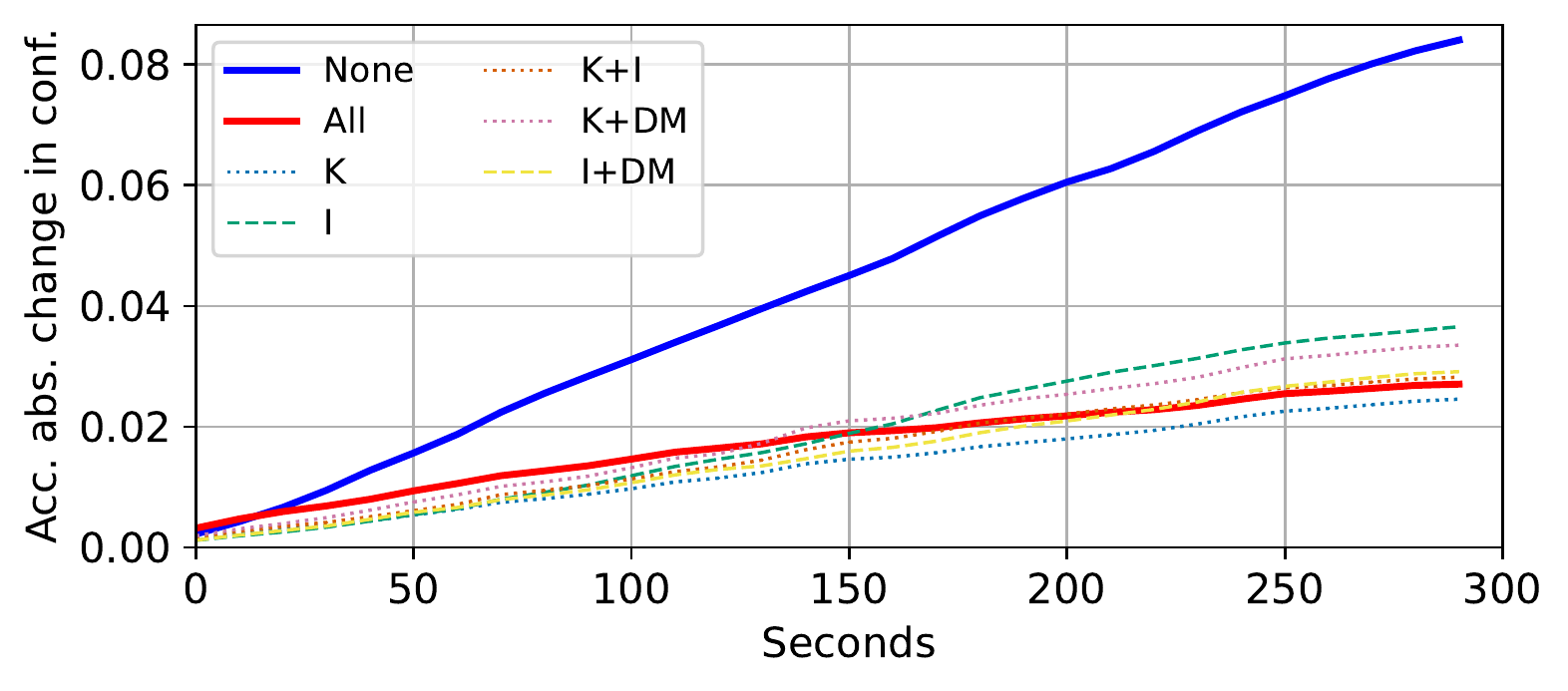}
        \vspace{-15pt}
        \caption{Cumulative absolute change in confidence.}
        \label{fig:cumulative_raw_score_change_absolute_time}
    \end{subfigure}
    \vspace{-5pt}
    \caption{The average change in model prediction over each edge as the interaction progresses in absolute time.
    The cause for the curve's dips is that not all interactions are of equal length. The further in time, the fewer interactions there are that have data at that time step. The abbreviations are S - Task Status, K - Task Knowledge, I - Task Intent, and DM - Dialogue Moves. \vspace{-8pt}}
    \label{fig:cumulative_changes_absolute_time}
\end{figure*}

As Task 2 is more approachable with more reasonable performance, we look further into the performance of the models as the interaction progresses. 
In this section, we look at not only the latent representation at the end of an interaction but also throughout the interaction, to predict the partner's missing knowledge.
We find that as the interaction progresses, in absolute time, there is an upward trend concerning both per edge F1 score (Figure \ref{fig:cumulative_per_edge_f1_absolute_time}), as well as concerning per task F1 score (Figure 
\ref{fig:cumulative_per_game_f1_absolute_time}).
This is not surprising. As models have access to richer interaction history, which enables them to make more reliable predictions of the partner's missing knowledge.
Furthermore, we observe that the augmented models improve the performance across time, confirming the results we obtained in the previous section.

We are also interested in the stability of system prediction over time, which is an important feature if these agents are ever employed in real interaction with humans. 
We introduce two metrics for this purpose: 
\begin{itemize}[leftmargin=*]
    \item \textbf{Number of Mind Changes}, which is the number of times that the model changes its prediction of whether an edge is missing from its partner's knowledge.
    \item \textbf{Accumulated Absolute Confidence Changes}, which adds up the absolute value of the changes in the prediction probability between timestamps.
\end{itemize}

Figure~\ref{fig:cumulative_mind_changes_absolute_time} shows the average number of mind changes and Figure~\ref{fig:cumulative_raw_score_change_absolute_time} shows the accumulated absolute confidence changes as the interaction progresses. 
We observe that the augmented models show significantly fewer changes in prediction and a lower change in prediction confidence, as compared with the base model. 
These results have shown that the base model is more inclined to ``change its mind'' in prediction as interaction proceeds, the augmented models that take into account of partner's mental states and dialogue moves are more stable in their prediction throughout the interaction.

%% file: src/5-conclusion.tex
\section{Discussion and Conclusion}
\label{sec:discussion}

In this work, we address the challenge of collaborative plan acquisition in human-agent collaboration.
We extend the \texttt{MindCraft} benchmark and formulate a problem for agents to predict missing task knowledge based on perceptual and dialogue history, focusing on understanding communicative intentions in Theory of Mind (ToM) modeling. 
Our empirical results highlight the importance of predicting the partner's missing knowledge and explicitly modeling their dialogue moves and mental states. 
A promising strategy for effective collaboration involves inferring and communicating missing knowledge to the partner while prompting them for their own missing knowledge. 
This collaborative approach holds the potential to improve decision-making when both agents actively engage in its implementation.
The findings have implications for the development of embodied AI agents capable of seamless collaboration and communication with humans. 
Specifically, by predicting their partner's missing knowledge and actively sharing that information, these agents can facilitate a shared understanding and successful execution of joint tasks. 
The future efforts following this research could explore and refine this collaborative strategy.

This work presents our initial results. It also has several limitations. 
The current setup assumes shared goals and a single optimal complete plan without alternatives, neglecting the complexity that arises from the absence of shared goals and the existence of alternative plans. 
Our motivation for controlling the form of partial plans and the predetermined complete plan is to enable a systematic focus on modeling and evaluating plan coordination behaviors. 
Although our current work is built on the \texttt{MindCraft} dataset where partial plans are represented by AND-graphs, 
the problem formulation can be potentially generalized to multiple AND-OR-graphs. 
Future research could explore approaches that incorporate AND-OR-graphs to account for alternative paths to achieving joint goals.

Additionally, our present study focuses on a dyadic scenario, employing human-human collaboration data to study collaborative plan acquisition. Since AI agents typically have limited visual perception and reasoning abilities compared to their human counterparts, the communication discourse is expected to exhibit increased instances of confirmations, repetitions, and corrections.
How to effectively extend the models trained on human-human data to human-agent collaboration remains an important question. With the emergence of large foundation models~\cite{bommasani2021opportunities}, our future work will incorporate these models into our framework to facilitate situated dialogue for collaborative plan acquisition. We will further conduct experiments and evaluate the efficacy of these models in more complex human-agent collaborations.

%% file: src/6-acknowledge.tex
\section{Acknowledgments}
This work was supported in part by NSF IIS-1949634 and NSF SES-2128623. 
The authors would like to thank the anonymous reviewers for their valuable feedback.

%% file: src/7-appendix.tex
\section{Appendix}
\label{sec:appendix}

\subsection{Dialogue Move Taxonomy and Annotation}
\label{sec:annotation}

\begin{figure}[htb!]
    \centering
    \includegraphics[width=0.95\columnwidth,trim={0cm 0cm 0cm 0cm},clip]{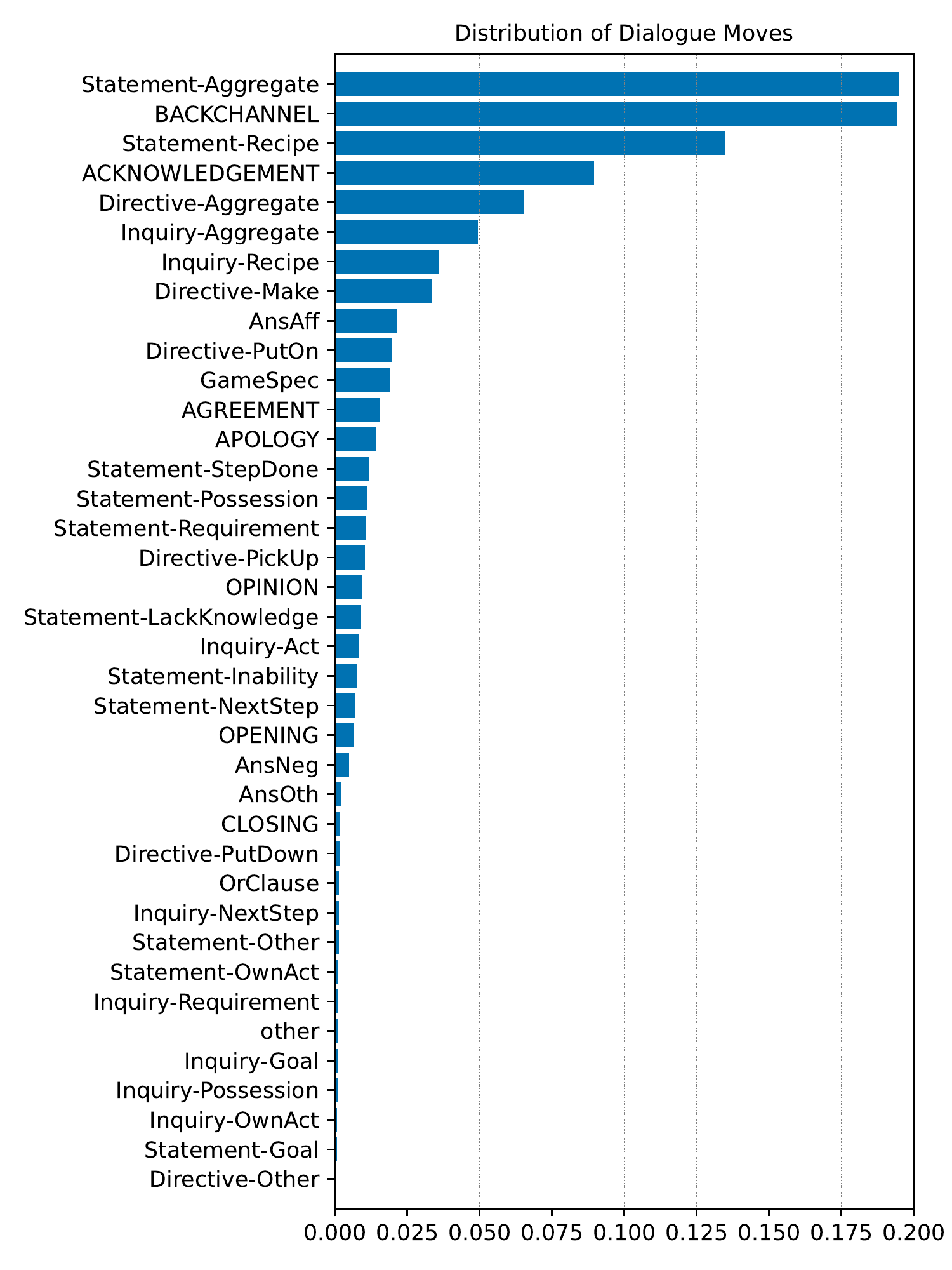}
    \vspace*{-10pt}
    \caption{Distribution of Dialogue Moves in our Dataset.\vspace{-8pt}}
    \label{fig:move_distr}
\end{figure}

The Dialogue Move consists of a dialogue category with up to three value slots which are representations of materials, mines, or tools used in the collaborative task. 
Table \ref{tab:dialogue_move_taxonomy} shows the full set of Dialogue moves with the type and number of parameters where applicable.
The distribution of the dialogue move categories is presented in Figure~\ref{fig:move_distr}.
The dialogue moves were labeled by a group of 7 human annotators. The dataset was divided into eight parts with one part being given to all annotators. The Cohen's kappa score, calculated on the common part of the dataset, between the annotators, if $k=0.807$ which is considered a strong agreement.

\vspace*{-10pt}

\subsection{Model Description}

\subsubsection{Input Modalities Processing}
Input processing of dialogue utterances, video frames, and plan representation is processed identically to \texttt{MindCraft}~\cite{bara-etal-2021-mindcraft}: dialogue utterances through BERT, video frames through a CNN, and partial plans through a GRU.

\subsubsection{Dialogue Moves Encoding} 
The dialogue move representations are groups of four one-hot embeddings representing the dialogue move type, see Table \ref{tab:dialogue_move_taxonomy}, followed three slot embeddings which are either a material, a tool, a mine, or to be ignored.

\subsubsection{Augmented Model} 
The augmented models utilize an LSTM as a sequence model to process the time series. On top of the three input modalities, there are four more optional inputs: a latent representation at every frame of when the partner's task status, task knowledge, task intention, or the dialogue move if one is available. 
The latent representations for the ToM tasks are retrieved from the output of the sequence model before the feed-forward networks.

\subsubsection{Computation Cost}
Our experiments were performed on an NVIDIA RTX A6000.
It takes around 60 min per experiment to train the model. 
At test time, it takes 82 ms on average per inference.

\subsection{Detailed Plan Prediction Results}

Table \ref{tab:complete_own_plan2} shows the change in performance for the task of predicting the player's own missing knowledge with all configurations of the augmented models.

\begin{table}[htb!]
    \centering
    \resizebox{\columnwidth}{!}{
        \begin{tabular}{|c|c|c|c|c|c|}
        \hline
        \textbf{Task} & \textbf{Tast} & \textbf{Task}. & \textbf{Dlg}. & \textbf{Per Edge} & \textbf{Avg. Per Game} \\
        \textbf{Status} & \textbf{Know}. & \textbf{Int.} & \textbf{Move} & \textbf{F1 Score} & \textbf{F1 Score} \\
        \hline
 
  &       &       &       &       17.0 $\pm$   0.2  &      19.8 $\pm$   1.0        \\ 
\texttt{X} &       &       &       &       20.4 $\pm$   1.1  &      23.3 $\pm$   1.5        \\ 
  
  &     \texttt{X} &       &       &       19.0 $\pm$   2.5  &      21.1 $\pm$   1.6        \\ 
\texttt{X} &     \texttt{X} &       &       &       20.9 $\pm$   1.5  &      23.3 $\pm$   2.2        \\ 
  
  &       &     \texttt{X} &       &       21.0 $\pm$   0.7  &      22.2 $\pm$   2.2        \\ 
\texttt{X} &       &     \texttt{X} &       &       19.4 $\pm$   1.3  &      23.5 $\pm$   1.3        \\ 
  
  &     \texttt{X} &     \texttt{X} &       &       19.6 $\pm$   1.4  &      21.4 $\pm$   1.7        \\ 
\texttt{X} &     \texttt{X} &     \texttt{X} &       &       18.2 $\pm$   1.0  &      21.5 $\pm$   2.4        \\ 
  
  &       &       &     \texttt{X} &       16.7 $\pm$   0.1  &      19.6 $\pm$   0.8        \\ 
\texttt{X} &       &       &     \texttt{X} &       20.4 $\pm$   1.4  &      22.6 $\pm$   0.8        \\ 
  
  &     \texttt{X} &       &     \texttt{X} &       20.1 $\pm$   1.4  &      22.1 $\pm$   1.2        \\ 
\texttt{X} &     \texttt{X} &       &     \texttt{X} &       20.9 $\pm$   1.2  &      24.3 $\pm$   2.1        \\ 
  
  &       &     \texttt{X} &     \texttt{X} &       19.8 $\pm$   1.7  &      21.7 $\pm$   1.8        \\ 
\texttt{X} &       &     \texttt{X} &     \texttt{X} &       19.8 $\pm$   0.8  &      24.0 $\pm$   2.9        \\ 
  
  &     \texttt{X} &     \texttt{X} &     \texttt{X} &       20.3 $\pm$   1.8  &      22.0 $\pm$   1.6        \\ 
\texttt{X} &     \texttt{X} &     \texttt{X} &     \texttt{X} &       17.4 $\pm$   0.1  &      20.0 $\pm$   1.9        \\ 
\hline
        \end{tabular}
    }
    \vspace*{-10pt}
    \caption{Change in performance in completing agent's own plan.}
    \label{tab:complete_own_plan2}
\end{table}

Table \ref{tab:complete_partner_plan2} shows the change in performance for the task of predicting the partner's missing knowledge with all augmenting input configurations.

\begin{table}[hb!]
    \centering
    \resizebox{\columnwidth}{!}{
        \begin{tabular}{|c|c|c|c|c|c|}
        \hline
        \textbf{Task} & \textbf{Task} & \textbf{Task}. & \textbf{Dlg}. & \textbf{Per Edge} & \textbf{Avg. Per Game} \\
        \textbf{Status} & \textbf{Know}. & \textbf{Int.} & \textbf{Move} & \textbf{F1 Score} & \textbf{F1 Score} \\
        \hline
 
  &       &       &       &       71.3 $\pm$   1.1  &      68.8 $\pm$   3.1  \\ 
\texttt{X} &       &       &       &       72.5 $\pm$   0.5  &      70.7 $\pm$   2.2  \\ 
  
  &     \texttt{X} &       &       &       74.4 $\pm$   0.3  &      73.6 $\pm$   1.4  \\ 
\texttt{X} &     \texttt{X} &       &       &       72.4 $\pm$   1.7  &      70.4 $\pm$   3.6  \\ 
  
  &       &     \texttt{X} &       &       74.5 $\pm$   1.4  &      73.8 $\pm$   3.0  \\ 
\texttt{X} &       &     \texttt{X} &       &       72.1 $\pm$   1.5  &      69.6 $\pm$   3.3  \\ 
  
  &     \texttt{X} &     \texttt{X} &       &       74.4 $\pm$   1.4  &      73.5 $\pm$   2.3  \\ 
\texttt{X} &     \texttt{X} &     \texttt{X} &       &       72.6 $\pm$   1.7  &      70.2 $\pm$   4.1  \\ 
  
  &       &       &     \texttt{X} &       71.4 $\pm$   1.0  &      69.1 $\pm$   3.1  \\ 
\texttt{X} &       &       &     \texttt{X} &       71.3 $\pm$   1.6  &      68.8 $\pm$   4.0  \\ 
  
  &     \texttt{X} &       &     \texttt{X} &       74.3 $\pm$   0.7  &      73.5 $\pm$   1.3  \\ 
\texttt{X} &     \texttt{X} &       &     \texttt{X} &       73.1 $\pm$   1.5  &      71.2 $\pm$   3.3  \\ 
  
  &       &     \texttt{X} &     \texttt{X} &       75.0 $\pm$   1.0  &      74.7 $\pm$   2.2  \\ 
\texttt{X} &       &     \texttt{X} &     \texttt{X} &       71.9 $\pm$   1.5  &      69.3 $\pm$   3.3  \\ 
  
  &     \texttt{X} &     \texttt{X} &     \texttt{X} &       73.4 $\pm$   1.2  &      72.4 $\pm$   2.0  \\ 
\texttt{X} &     \texttt{X} &     \texttt{X} &     \texttt{X} &       73.5 $\pm$   0.5  &      72.1 $\pm$   1.8  \\ 
\hline
        \end{tabular}
    }
    \vspace*{-10pt}
    \caption{Change in performance on completing the partner's plan.}
    \label{tab:complete_partner_plan2}
\end{table}

\begin{table*}[htb!]
    \centering
    \resizebox{\textwidth}{!}{
        \begin{tabular}{|l|l|l|}
        \hline
        \textbf{Name} & \textbf{Parameters} & \textbf{Description or Example} \\
                \hline
        Directive-Make & Block &  Directing their partner to make a material\\

        Directive-Other & N/A & Ambiguous directive\\
 
        Directive-PickUp & Block & Directing their partner to use their tool on a block\\

        Directive-PutDown & [Block] & Directing partner to place the block on the ground\\
 
        Directive-PutOn & Block & Directing partner to place block on anther block\\

        Inquiry-Act & N/A & Asking what their partner is doing \\
 
        Inquiry-Goal & [Block] & Inquiring about partner's goal \\

        Inquiry-NextStep & N/A & Asking what should be done next \\
 
        Inquiry-OwnAct & [Block or Block-Pair] & Asking what they should do next\\

        Inquiry-Possession & N/A & Asking about partner's tools\\
 
        Inquiry-Recipe & Block & Asking about a particular recipe node\\

        Inquiry-Requirement & Block & Asking about required tools or blocks\\
 
        Statement-Goal & Block & Stating their goal\\

        Statement-Inability & N/A & Stating an inability \\
 
        Statement-LackKnowledge & [Block] & Stating their lack of knowledge about a block\\

        Statement-NextStep & Block or Block-Pair & Stating their next step\\
 
        Statement-Other & N/A & Ambiguous statement\\

        Statement-OwnAct & Block & Statement about the player's own current act\\
 
        Statement-Possession & Tool & Statement about the player's own inventory \\

        Statement-Recipe & Block and Block-Pair & Statement describing a step in the recipe \\
 
        Statement-Requirement & Tool or Block or Pair & Statement about required tools or blocks for a step \\

        Statement-StepDone & Block & Statement informing about completion of a step \\
 
        BACKCHANNEL         & N/A & off topic statements          \\

        OPINION             & N/A & We should ...; it must be ... \\

        AGREEMENT                 & N/A            & sure; ok;                                      \\

        AnsAff              & N/A & agreement to a question       \\
 
        CLOSING                 & N/A              & bye; i'm out                                   \\

        AnsNeg              & N/A & nope, i don't have           \\
 
        ACKNOWLEDGMENT      & N/A & right; done; indeed          \\

        AnsOth                    & N/A            & I don't know                                   \\
 
        OPENING                              & N/A & hi; hello                                      \\

        OrClause            & N/A                  & ... or ...                                     \\
 
        APOLOGY             & N/A                  & sorry                                          \\

        GameSpec             & N/A                 & Specific to the game environment        \\
 
        other               & N/A & \\
        \hline
        \end{tabular}
    }
    \caption{The Dialogue move taxonomy. The information portraying dialogue acts was expanded to give a more fine-grained description.}
    \label{tab:dialogue_move_taxonomy}
\end{table*}